\documentclass[conference]{IEEEtran}
\IEEEoverridecommandlockouts
\usepackage{cite}
\usepackage{amsmath,amssymb,amsfonts}
\usepackage{algorithmic}
\usepackage{graphicx}
\usepackage{textcomp}
\usepackage{xcolor}
\usepackage{booktabs}
\usepackage{multirow}
\usepackage{url}
\usepackage{hyperref}
\hypersetup{hidelinks}

\def\BibTeX{{\rm B\kern-.05em{\sc i\kern-.025em b}\kern-.08em
    T\kern-.1667em\lower.7ex\hbox{E}\kern-.125emX}}

\begin{document}

\title{Trimodal Deep Learning for Glioma Survival Prediction: A Feasibility Study Integrating Histopathology, Gene Expression, and MRI}

\author{\IEEEauthorblockN{Iain Swift}
\IEEEauthorblockA{\textit{Department of Computer Science} \\
\textit{Munster Technological University}\\
Cork, Ireland \\
iain.swift@mymtu.ie}
\and
\IEEEauthorblockN{Jing Hua Ye}
\IEEEauthorblockA{\textit{Department of Computer Science} \\
\textit{Munster Technological University}\\
Cork, Ireland} 
JingHua.Ye@mtu.ie}

\maketitle

\begin{abstract}
Multimodal deep learning has improved prognostic accuracy for brain tumours by integrating histopathology and genomic data, yet the contribution of volumetric MRI within unified survival frameworks remains unexplored. This pilot study extends a bimodal framework by incorporating Fluid Attenuated Inversion Recovery (FLAIR) MRI from BraTS2021 as a third modality. Using the TCGA-GBMLGG cohort (664 patients), we evaluate three unimodal models, nine bimodal configurations, and three trimodal configurations across early, late, and joint fusion strategies. In this small cohort setting, trimodal early fusion achieves an exploratory Composite Score (CS\,=\,0.854), with a controlled $\Delta$CS of +0.011 over the bimodal baseline on identical patients, though this difference is not statistically significant ($p$\,=\,0.250, permutation test). MRI achieves reasonable unimodal discrimination (CS\,=\,0.755) but does not substantially improve bimodal pairs, while providing measurable uplift in the three-way combination. All MRI containing experiments are constrained to 19 test patients, yielding wide bootstrap confidence intervals (e.g.\ [0.400,\,1.000]) that preclude definitive conclusions. These findings provide preliminary evidence that a third imaging modality may add prognostic value even with limited sample sizes, and that additional modalities require sufficient multimodal context to contribute effectively.
\end{abstract}


% ============================================================
\section{Introduction}
% ============================================================

Gliomas are the most prevalent malignant primary brain tumours in adults, accounting for approximately 80\% of all malignant primary brain tumours~\cite{hanif2017}. Prognosis varies dramatically: glioblastoma multiforme (GBM) typically has a median survival under 15 months despite maximal treatment, while lower-grade gliomas (LGG) show more variable progression~\cite{bakas2018brats}. Modern classification integrates molecular markers such as isocitrate dehydrogenase (IDH) mutation status and 1p/19q codeletion with traditional histopathological grading, yet accurate survival prediction remains challenging due to marked genetic and cellular heterogeneity among patients with similar clinical grades~\cite{tcga2008}.

Each clinical data modality captures a distinct aspect of tumour biology. RNA sequencing (RNA-seq) reveals transcriptomic activity and pathway dysregulation~\cite{steyaert2023}. Whole-slide imaging (WSI) of haematoxylin and eosin (H\&E)-stained tissue captures cellular morphology, necrosis patterns, and vascular proliferation. Magnetic resonance imaging (MRI) provides volumetric information on tumour extent, peritumoural oedema, and enhancement patterns that complement histopathology and genomics~\cite{lee2023mri,nie2019mri}. Single modality models capture only partial views of this biology, motivating multimodal integration.

Steyaert et al.~\cite{steyaert2023} demonstrated a multimodal deep learning framework integrating WSI and RNA-seq for glioma survival prediction, achieving a Composite Score (CS) of 0.836 using early fusion on the TCGA-GBMLGG cohort. They noted in their conclusion that the framework could be expanded to include radiographic images such as MRI. However, two key questions remain unanswered: (1)~whether adding MRI as a third modality provides additional prognostic value beyond bimodal integration, and (2)~how MRI's contribution depends on the presence of other modalities and the fusion strategy employed.

This paper makes two contributions. First, we extend the bimodal framework to trimodal integration by incorporating FLAIR MRI sequences from BraTS2021~\cite{bakas2018brats}, implementing a 3D ResNet-18 encoder for volumetric feature extraction. Second, we systematically evaluate all unimodal, bimodal, and trimodal configurations across early, late, and joint fusion strategies, including a controlled comparison on identical patient sets to isolate MRI's contribution and a supplementary analysis of attention-based fusion mechanisms (bilinear, cross attention, gated) for bimodal histopathology genomics integration.

% ============================================================
\section{Related Work}
% ============================================================

Cox proportional hazards (Cox-PH) models remain the standard for survival analysis with censored data~\cite{cox1972}. DeepSurv~\cite{katzman2018deepsurv} and Cox-nnet~\cite{ching2018coxnnet} demonstrated that neural extensions of Cox-PH outperform traditional models. For brain tumours, Mobadersany et al.~\cite{mobadersany2018} combined histopathology CNNs with genomic markers, Chen et al.~\cite{chen2022pathomic} proposed pathomic fusion using attention gating, and Steyaert et al.~\cite{steyaert2023} implemented early, late, and joint fusion of WSI and RNA-seq, finding early fusion superior.

MRI-based survival prediction has been explored independently: Lee et al.~\cite{lee2023mri} demonstrated prognostic value from 3D deep learning features, and Nie et al.~\cite{nie2019mri} applied multi-channel 3D feature learning. More recently, Chen et al.~\cite{chen2021mcat} proposed multimodal co-attention (MCAT) for WSI-genomic fusion, and Jaume et al.~\cite{jaume2024survpath} introduced hierarchical token-based modelling in SurvPath. Pathology foundation models such as UNI~\cite{chen2024uni} and CTransPath~\cite{wang2022ctranspath} now provide general-purpose WSI encoders that may replace ImageNet pretrained backbones; we adopt simpler encoders here to isolate the effect of adding MRI as a third modality. However, systematic integration of histopathology, gene expression, and MRI within a unified framework has not been demonstrated. Three fusion strategies, early (feature concatenation), late (score combination), and joint (end-to-end training), are commonly employed, but whether early fusion's bimodal superiority extends to trimodal integration is unknown.

% ============================================================
\section{Materials and Methods}
% ============================================================

\subsection{Dataset and Cohort}
Brain tumour data were sourced from the TCGA-GBMLGG cohort~\cite{tcga2008} and the BraTS2021 challenge~\cite{bakas2018brats}. Steyaert et al.~\cite{steyaert2023} reported 783 patients, including both adult and paediatric cases; our cohort of 664 adult patients (426 LGG, 238 GBM) excludes the paediatric subset and additionally removes patients lacking survival times or matched clinical metadata (diagnosis, vital status). Of these, 268 (40.4\%) had recorded death events, and 396 (59.6\%) were censored. Median survival was 18.9 months.

Three data modalities were used: (i) formalin-fixed paraffin-embedded (FFPE) whole-slide H\&E images for 590 patients, (ii) gene expression profiles (RNA-seq and microarray) for 509 patients covering 12,778 genes after preprocessing, and (iii) FLAIR MRI volumes from BraTS2021 mapped to TCGA identifiers for 162 patients. All 162 MRI-matched patients were identified through exhaustive linkage of BraTS2021 subject identifiers to TCGA case IDs; no additional FLAIR volumes were available for this cohort.

The key constraint is MRI availability. Modality intersections determine feasible experiments: FFPE+RNA experiments use up to 490 patients, while all MRI containing experiments are limited to 47--77 patients for training with 19 for testing. Table~\ref{tab:splits} summarises data splits with GBM proportions that affect model training.

\begin{table}[t]
\caption{Data Splits by Modality Combination}
\label{tab:splits}
\centering
\small
\begin{tabular}{lccc}
\toprule
\textbf{Experiment} & \textbf{Train} & \textbf{Test} & \textbf{GBM\%} \\
\midrule
FFPE Only        & 465 & 117 & 36.2\%$^{\S}$ \\
RNA Only         & 389 & 120 & 34.2\%$^{\S}$ \\
MRI Only         & 143 &  19 & 68.5\%        \\
FFPE + RNA       & 376 & 114 & 15.7\%        \\
RNA + MRI        &  47 &  19 & 25.5\%        \\
FFPE + MRI       &  58 &  19 & 15.8\%        \\
FFPE+RNA+MRI     &  47 &  19 & 25.5\%        \\
\bottomrule
\multicolumn{4}{l}{\footnotesize $^{\S}$Approx.; reflects modality-available subset (35.8\% GBM).}
\end{tabular}
\end{table}

\subsection{Data Preprocessing}

\subsubsection{Gene Expression}
Genes with missing values were removed. Counts were log-transformed, followed by z-score normalisation. ComBat-Seq~\cite{zhang2020combat} corrected batch effects between TCGA RNA-seq and microarray data. The final expression vector contained 12,778 genes per sample.

\subsubsection{Histopathology}
FFPE whole-slide images were segmented using OTSU thresholding on HSV-converted tissue to separate foreground from background. Non-overlapping patches of $224 \times 224$ pixels were extracted at $20\times$ magnification using OpenSlide, with up to 4,000 patches per slide. Stain augmentation was applied using random ColorJitter (PyTorch). Patches were normalised using ImageNet statistics (mean=[0.485, 0.456, 0.406], std=[0.229, 0.224, 0.225]).

\subsubsection{MRI (FLAIR Volumes)}
BraTS2021 FLAIR volumes were mapped to TCGA patient identifiers. Preprocessing comprised skull stripping using BraTS-provided brain masks, intensity normalisation via percentile clipping (1st--99th percentile, rescaled to $[0,1]$), resampling to $1\,\text{mm}^3$ isotropic voxel spacing with trilinear interpolation, and centre cropping/padding to $128 \times 128 \times 128$ voxels. Training augmentation included random 3D flips ($p$\,=\,0.5 per axis), random rotation ($p$\,=\,0.5), random cropping (fraction\,=\,0.85), Gaussian noise ($\sigma$\,=\,0.01), and random intensity shifts (range\,=\,0.1).

\subsection{Unimodal Models}

\subsubsection{Histopathology Model}
A ResNet-50 CNN~\cite{he2016resnet} pretrained on ImageNet was used, with only the final ResNet block fine-tuned to prevent overfitting. Each $224 \times 224$ patch produces a 2048-dimensional feature vector. During training, 100 random patches per WSI are used, with patch-level Cox risk scores predicted. During inference, risk scores are averaged across all patches per patient.

\subsubsection{Gene Expression Model}
A three-layer MLP maps 12,778 gene expression values through hidden layers of 4,096 and 2,048 units with dropout (0.5) and ReLU activations, outputting a 2048-dimensional feature vector mapped to a single Cox risk score.

\subsubsection{MRI Model}
A 3D ResNet-18 processes single-channel FLAIR volumes ($128^3$). The architecture uses 3D convolutions with BasicBlock3D modules, adaptive average pooling to a 512-dimensional vector, followed by a two-stage projection head ($512 \rightarrow 512 \rightarrow 2048$ with batch normalisation, ReLU, and dropout) for fusion compatibility. ResNet-18 was chosen over ResNet-50 due to the small MRI training set ($n$\,=\,143) to reduce overfitting risk. Standard ImageNet pretraining does not apply to 3D architectures; while 3D pretrained medical encoders exist (e.g.\ Med3D~\cite{chen2019med3d}, Models Genesis~\cite{zhou2021models}), we train from scratch to maintain a controlled comparison and avoid confounds from domain-adapted weight transfer. Training therefore proceeds from random initialisation with separate learning rates for backbone ($1\text{e-}5$) and head ($1\text{e-}4$).

\subsection{Multimodal Fusion Strategies}

Two fusion strategies from Steyaert et al.~\cite{steyaert2023} were reimplemented, with joint fusion additionally evaluated.

\subsubsection{Early Fusion}
Unimodal feature vectors are concatenated (4096-dim for bimodal, 6144-dim for trimodal) and fed into a fusion MLP. For MRI-containing experiments, the fusion MLP uses reduced hidden dimensions ($512$) with dropout ($0.3$) due to smaller training sets.

\subsubsection{Late Fusion}
Independent unimodal models produce separate risk scores, which are combined via Cox proportional hazards regression with cross-validation. This approach is architecturally the simplest and most robust to missing modalities.

\subsubsection{Joint Fusion}
Joint fusion simultaneously trains all modality encoders end-to-end with a shared Cox loss, allowing gradient flow from the survival objective through each encoder. For FFPE+RNA, the ResNet-50 backbone (final two layers fine-tuned) and RNA MLP are optimised jointly; feature vectors are concatenated and processed by a fusion MLP producing a single risk score. For MRI-containing combinations, the 3D ResNet-18 encoder is added with a separate learning rate. Unlike early fusion, which relies on features extracted independently, joint fusion enables the encoders to learn representations optimised for the multimodal survival task.

\subsection{Attention-Based Fusion}

As a supplementary analysis, three attention mechanisms were tested on FFPE+RNA early fusion to assess whether more expressive architectures improve discrimination: bilinear fusion ($y = x_{\text{WSI}}^T W x_{\text{RNA}}$), cross-attention ($\text{Attn}(Q\!=\!\text{WSI}, K\!=\!\text{RNA}, V\!=\!\text{RNA})$), and gated fusion ($y = g(x_{\text{WSI}}) \cdot x_{\text{RNA}} + (1\!-\!g) \cdot x_{\text{WSI}}$). All models used lr\,=\,0.0001, hidden dimension\,=\,256, dropout\,=\,0.25, and were trained for 80 epochs.

\subsection{Loss Function and Evaluation}

\begin{figure}[t]
\centering
\includegraphics[width=\columnwidth]{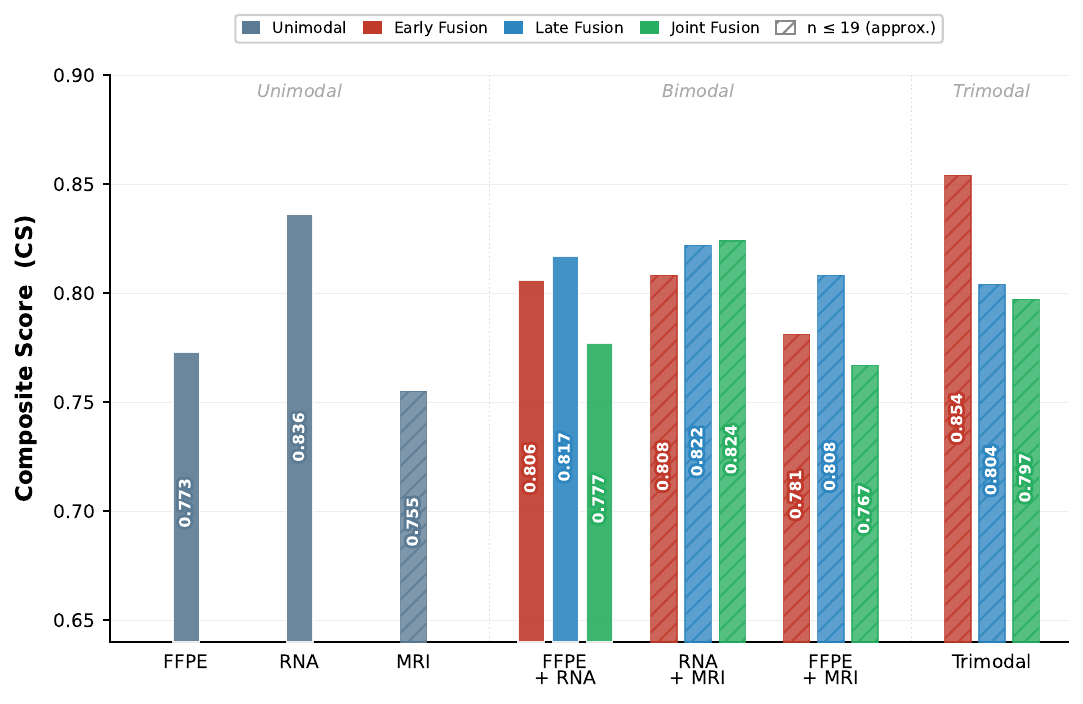}
\caption{Composite Score (CS) across all configurations, grouped by modality combination and fusion strategy. Hatched bars (\textsuperscript{$\ddagger$}) indicate experiments with $n$\,$\leq$\,19 test patients where IBS is imprecise. Trimodal early fusion achieves the highest CS (0.854).}
\label{fig:cs_barchart}
\end{figure}

All models are trained with the Cox proportional hazards partial likelihood loss~\cite{cox1972}. Performance is evaluated using the Composite Score CS\,=\,(CI\,+\,(1$-$IBS))/2, which averages the concordance index (CI; 1.0\,=\,perfect, 0.5\,=\,random)~\cite{harrell1982} with the complement of the Integrated Brier Score (IBS; 0\,=\,perfect calibration). Bootstrap 95\% confidence intervals (10,000 resamples) are computed over CI. For experiments with $n$\,$\leq$\,25 test patients, IBS estimates are imprecise, and the resulting CS values should be treated as approximate. A grouped bar chart (Fig.~\ref{fig:cs_barchart}) provides visual comparison across all configurations.

% ============================================================
\section{Experimental Configuration}
% ============================================================

All models used Adam optimisation with early stopping on validation CI. FFPE+RNA experiments used a 10-fold CV with 114 held-out test patients; MRI-containing experiments used a 5-fold CV with 19 test patients. Joint fusion used per-encoder learning rates to balance gradient magnitudes across modalities. All training was performed on an NVIDIA RTX 3060 (12\,GB). To isolate MRI's contribution from cohort-difference confounds, a controlled experiment trained FFPE+RNA early fusion on the identical 47/19 patient split used for trimodal experiments. The difference in cross-validation folds (10-fold vs.\ 5-fold) and test-set sizes (114 vs.\ 19) across experiments reflects sample-size constraints imposed by MRI availability rather than a methodological inconsistency; in all cases, we maximised the number of folds while retaining stable per-fold training sets. Bootstrap confidence intervals are reported where test-set predictions were available; missing entries in Table~\ref{tab:results} reflect single-split evaluations for which bootstrapping was not performed.

% ============================================================
\section{Results}
% ============================================================

\subsection{Unimodal Performance}

Table~\ref{tab:results} presents the complete results. Among unimodal models, RNA expression achieves the strongest performance (CS\,=\,0.836), followed by FFPE histopathology (CS\,$\approx$\,0.773) and MRI (CS\,=\,0.755). Despite the small MRI training set ($n$\,=\,143) with 68.5\% GBM, a GBM-dominated subtype distribution, and the absence of pretrained weights for 3D convolutions, the MRI encoder achieves meaningful discrimination comparable to histopathology.

\begin{table}[t]
\caption{Composite Score (CS) Across All Configurations}
\label{tab:results}
\centering
\begin{tabular}{llc}
\toprule
\textbf{Model} & \textbf{Fusion} & \textbf{Test CS} \\
\midrule
\multicolumn{3}{l}{\textit{Unimodal}} \\
FFPE & --- & 0.773 \\
RNA & --- & 0.836 \\
MRI & --- & 0.755$^{\ddagger}$ \\
\midrule
\multicolumn{3}{l}{\textit{Bimodal}} \\
FFPE + RNA & Early & 0.806 \\
FFPE + RNA & Late & 0.817 \\
FFPE + RNA & Joint & 0.777 \\
RNA + MRI & Early & 0.808$^{\ddagger}$ \\
RNA + MRI & Late & 0.822$^{\ddagger}$ \\
RNA + MRI & Joint & 0.824$^{\ddagger}$ \\
FFPE + MRI & Early & 0.781$^{\ddagger}$ \\
FFPE + MRI & Late & 0.808$^{\ddagger}$ \\
FFPE + MRI & Joint & 0.767$^{\ddagger}$ \\
\midrule
\multicolumn{3}{l}{\textit{Trimodal}} \\
FFPE+RNA+MRI & Early & \textbf{0.854}$^{\ddagger}$ \\
FFPE+RNA+MRI & Late & 0.804$^{\ddagger}$ \\
FFPE+RNA+MRI & Joint & 0.797$^{\ddagger}$ \\
\midrule
\multicolumn{3}{l}{\textit{Controlled Comparison}} \\
FFPE+RNA$^{\dagger}$ & Early & 0.843$^{\ddagger}$ \\
\bottomrule
\multicolumn{3}{l}{\footnotesize $^{\dagger}$Restricted to same 47 train / 19 test patients as trimodal.}\\
\multicolumn{3}{l}{\footnotesize $^{\ddagger}$CS approximate ($n$\,$\leq$\,19 test); IBS component imprecise.}
\end{tabular}
\end{table}

\subsection{Bimodal Fusion}

FFPE+RNA fusion achieves CS\,=\,0.806 (early) and CS\,=\,0.817 (late), consistent with Steyaert et al.'s~\cite{steyaert2023} reported CS of 0.836. The bimodal FFPE+RNA late fusion Cox LASSO regression recovered equal weights ($\beta_{\text{path}} = 0.5$, $\beta_{\text{RNA}} = 0.5$), indicating that neither unimodal risk score dominated when combined at the score level. RNA+MRI achieves CS\,=\,0.808 (early) and CS\,=\,0.822 (late), demonstrating that RNA's strong signal carries through even with the small MRI-matched cohort. FFPE+MRI per-fold CI analysis reveals near-chance discrimination (mean CI\,$\approx$\,0.50); the CS values in Table~\ref{tab:results} (0.767--0.808) are elevated by favourable IBS estimates on $n$\,=\,19 patients and should be interpreted cautiously. Fig.~\ref{fig:cs_barchart} visualises the performance landscape across all configurations and fusion strategies.

\subsection{Joint Fusion}

FFPE+RNA joint fusion achieves CS\,=\,0.777 (10-fold CV), lower than both early (0.806) and late (0.817) fusion. End-to-end fine-tuning of the ResNet-50 backbone on 376 training samples increases overfitting risk compared to using pre-extracted frozen features.

Among MRI-containing experiments, RNA+MRI joint fusion achieves CS\,=\,0.824 (bootstrap 95\% CI: [0.500, 1.000]), marginally above early (0.808) and late (0.822) fusion on the same test set. This suggests that end-to-end gradient flow enables the MRI encoder to learn more informative representations when guided by RNA's strong supervisory signal, though the wide confidence interval precludes definitive conclusions. FFPE+MRI joint fusion (CS\,=\,0.767) is the lowest among FFPE+MRI variants, consistent with the near-chance per-fold CI for this combination. Trimodal joint fusion (0.797) falls below both early (0.854) and late (0.804), ranking as: early $>$ late $>$ joint.

\subsection{Trimodal Fusion and Controlled Comparison}

Three-way early fusion (FFPE+RNA+MRI) achieves the highest CS of 0.854 across all experiments. The na\"ive comparison with the full-cohort FFPE+RNA baseline yields $\Delta$CS\,=\,+0.048 (0.854 vs.\ 0.806), but this comparison confounds MRI's contribution with cohort differences. The controlled comparison, using the identical 47-patient training set and 19-patient test set, yields FFPE+RNA CS\,=\,0.843 and trimodal CS\,=\,0.854, giving a \textbf{controlled $\Delta$CS\,=\,+0.011} that isolates MRI's contribution on identical patients.

Per-fold CI ranges from 0.773 to 0.932 (mean $\pm$ std: 0.845 $\pm$ 0.051), reflecting sensitivity to the small test set ($n$\,=\,19). A permutation test comparing trimodal to the restricted bimodal across folds yielded $p$\,=\,0.250, which is not significant; however, the minimum achievable $p$-value with 5-fold paired permutation is $1/2^5 = 0.031$, so this test has inherently limited resolving power.

Three-way late fusion achieves CS\,=\,0.804, with equal-weight combination ($0.333$ per modality). The bimodal late fusion similarly recovered equal weights ($0.5$ per modality), but with only two informative inputs, equal weighting was less harmful than diluting across three modalities, including MRI's less discriminative scores.

Table~\ref{tab:mri_impact} summarises the impact of adding MRI to existing baselines.

\begin{table}[t]
\caption{Impact of Adding MRI (Early Fusion). The controlled $\Delta$CS comparison uses identical patients to isolate MRI's contribution.}
\label{tab:mri_impact}
\centering
\begin{tabular}{lccc}
\toprule
\textbf{Baseline} & \textbf{+ MRI} & \textbf{$\Delta$CS} & \textbf{Note} \\
\midrule
FFPE (0.773) & FFPE+MRI (0.781) & +0.008 & Uncontrolled$^{\ddagger}$ \\
RNA (0.836) & RNA+MRI (0.808) & $-$0.028 & Uncontrolled$^{\ddagger}$ \\
FFPE+RNA (0.806) & Trimodal (0.854) & +0.048 & Uncontrolled \\
FFPE+RNA$^{\dagger}$ (0.843) & Trimodal (0.854) & \textbf{+0.011} & Controlled \\
\bottomrule
\multicolumn{4}{l}{\footnotesize $^{\dagger}$Restricted to same 47/19 patient split as trimodal.}\\
\multicolumn{4}{l}{\footnotesize $^{\ddagger}$Different patient subsets and sample sizes; $\Delta$CS confounded}\\
\multicolumn{4}{l}{\footnotesize by cohort composition differences (see Table~\ref{tab:splits}).}
\end{tabular}
\end{table}

\subsection{Attention-Based Fusion}

As a supplementary analysis, Table~\ref{tab:attention} presents results for attention mechanisms on FFPE+RNA early fusion.

\begin{table}[t]
\caption{Attention-Based Fusion Models (FFPE + RNA Early Fusion)}
\label{tab:attention}
\centering
\begin{tabular}{lccc}
\toprule
\textbf{Model} & \textbf{Test CI} & \textbf{Parameters} & \textbf{Train CI} \\
\midrule
Bilinear & \textbf{0.819} & 537K & 0.979 \\
Cross-Attention & 0.814 & 1.84M & 0.975 \\
Gated & 0.807 & 3.18M & 0.954 \\
Baseline MLP & 0.801 & $\sim$4.2M & $\sim$0.95 \\
\bottomrule
\end{tabular}
\end{table}

Bilinear fusion achieves the best test CI (0.819), outperforming the baseline MLP by +0.018. All attention models improve over the baseline, but gains are modest. The train-test gap (bilinear: 0.979 vs.\ 0.819) indicates some overfitting. Gated fusion, with the most parameters (3.18M), achieves the lowest test performance among attention models, suggesting over-parameterisation for this dataset size.

% ============================================================
\section{Discussion}
% ============================================================

\subsection{MRI Contribution}

MRI achieves meaningful unimodal discrimination (CS\,=\,0.755), comparable to histopathology (0.773) and well above chance. Given only 143 training patients, a GBM-heavy distribution, and no dedicated 3D pretraining, this score likely reflects a data-limited performance ceiling rather than an inherent upper bound on FLAIR's prognostic value. The controlled trimodal uplift ($\Delta$CS\,=\,+0.011 on identical patients) suggests that MRI captures complementary structural information, though the small magnitude and imprecise IBS on $n$\,=\,19 test patients preclude strong claims. Adding MRI to RNA yields a modest CS drop (RNA+MRI\,=\,0.808 vs.\ RNA\,=\,0.836, $\Delta$CS\,=\,$-$0.028); FFPE+MRI per-fold CI averages near chance ($\approx$0.50), and the elevated CS values for FFPE+MRI (0.767--0.808) stem from the IBS component on small samples rather than meaningful discrimination. Because these bimodal comparisons use different patient subsets and substantially smaller training sets than the unimodal baselines (Table~\ref{tab:splits}), $\Delta$CS values for uncontrolled rows should be interpreted cautiously.

The pattern---neutral-to-negative in bimodal, positive in trimodal---suggests that MRI's structural signal becomes beneficial only when combined with both histopathological and transcriptomic context. With genomic data anchoring the prediction, MRI contributes complementary volumetric information; without it, MRI's overlapping visual features add little to histopathology alone.

The GBM-dominated MRI training set (68.5\% GBM vs.\ 15.7\% in FFPE+RNA) may still limit MRI performance relative to its potential by reducing survival variability. Notably, the trimodal training set (47 patients, 25.5\% GBM) has a more balanced GBM/LGG ratio than the MRI-only set (68.5\% GBM), which may itself contribute to improved discrimination by restoring survival variability. The controlled comparison mitigates this partially---bimodal and trimodal models train on identical patients---but the possibility that the trimodal improvement is partly explained by the favourable composition of the MRI-intersected subset, rather than MRI's informational content per se, cannot be fully excluded.

These findings align with Steyaert et al.'s~\cite{steyaert2023} suggestion that radiographic images may improve survival prediction, while highlighting that benefit requires adequate multimodal context and sample size. Taken together, the results support the feasibility of incorporating MRI into a trimodal framework, but do not yet establish its definitive benefit.

\subsection{Fusion Strategy Comparison}

Early fusion outperforms late fusion for trimodal integration (0.854 vs.\ 0.804), consistent with Steyaert et al.~\cite{steyaert2023}. Trimodal late fusion's equal weights ($0.333$ per modality) dilute RNA's strong signal across three inputs, whereas early fusion implicitly weights contributions through learned feature representations. The equal weights recovered by the LASSO-regularised Cox model suggest that the cross-validated penalty parameter favoured simpler, near-equal weighting when training data were limited, and modality scores had similar marginal scale. Joint fusion reveals dataset-size sensitivity: for FFPE+RNA ($n$\,=\,376), joint underperforms early and late (0.777 vs.\ 0.806/0.817) due to end-to-end overfitting, but for RNA+MRI ($n$\,=\,47), joint achieves the highest bimodal CS (0.824), suggesting gradient flow benefits smaller experiments anchored by a strong co-modality. Bilinear attention achieves the best bimodal FFPE+RNA CI (0.819) with the fewest parameters (537K), indicating compact multiplicative architectures outperform higher-capacity models on this dataset.

\subsection{Comparison and Clinical Context}

Our FFPE+RNA Composite Scores (early CS\,=\,0.806, late CS\,=\,0.817) approach Steyaert et al.'s~\cite{steyaert2023} reported CS of 0.836, with the gap likely attributable to our smaller effective cohort (490 vs.\ 783 patients) and the use of an ImageNet-pretrained ResNet-50 rather than a pathology-specific encoder. Notably, RNA alone achieves CS\,=\,0.836, matching Steyaert et al.'s multimodal result and suggesting that gene expression dominates the prognostic signal. Since MRI is routinely acquired for all glioma patients, a trimodal model leveraging all available modalities is clinically appealing; bimodal FFPE+RNA fusion remains effective when MRI is unavailable.

% ============================================================
\section{Limitations}
% ============================================================

The primary limitation is the small MRI cohort: only 162 TCGA patients have BraTS2021 FLAIR volumes, yielding 19 test patients for all MRI experiments with high-variance CI estimates and insufficient power for log-rank significance testing. Only single-channel FLAIR was used; multi-sequence integration and self-supervised 3D pretraining could improve MRI performance. The TCGA cohort over-represents surgical candidates, as tissue collection requires surgical resection, excluding patients with unresectable tumours. Additionally, stratification by histological diagnosis rather than IDH mutation status limits alignment with the 2021 WHO classification. Composite Score estimates for MRI-containing experiments ($n$\,$\leq$\,19) rely on imprecise IBS values; time-dependent AUC would provide an additional discrimination metric. A clinical baseline model using age, grade, and IDH mutation status was not evaluated, which limits contextualisation of the deep learning results against standard clinical predictors. No external validation cohort with matched WSI, RNA-seq, and MRI currently exists.

% ============================================================
\section{Conclusion}
% ============================================================

This pilot study extends bimodal glioma survival prediction to trimodal integration by incorporating FLAIR MRI from BraTS2021 as a third data modality. MRI achieves meaningful unimodal discrimination (CS\,=\,0.755), and trimodal early fusion (FFPE+RNA+MRI) achieves the highest observed Composite Score (0.854) across all experiments, with a controlled comparison yielding $\Delta$CS\,=\,+0.011 over the bimodal baseline on identical patients. This difference is not statistically significant ($p$\,=\,0.250, permutation test) and the bootstrap 95\% CI [0.400,\,1.000] is wide, so the result should be interpreted as preliminary evidence rather than a confirmed finding. Nonetheless, the consistent direction of improvement across fusion strategies suggests that complementary structural information may add prognostic value even with severely limited sample sizes.

MRI's contribution is context-dependent: it does not substantially improve bimodal pairs in early/late fusion (FFPE+MRI per-fold CI\,$\approx$\,0.50, RNA+MRI CS\,=\,0.808) yet provides uplift in the trimodal setting (CS\,=\,0.854), suggesting that complementary value emerges when both histopathological and transcriptomic context are present. Joint fusion reveals a complementary pattern: end-to-end training benefits MRI most when paired with RNA (joint CS\,=\,0.824 vs.\ early CS\,=\,0.808), but increases overfitting risk for larger histopathology-dominant cohorts. Attention-based fusion provides modest but consistent gains for bimodal fusion, with bilinear fusion achieving the best performance (CI\,=\,0.819) using the fewest parameters.

These results motivate future work in three directions: (1) larger MRI-matched cohorts to establish MRI's full contribution with adequate statistical power, (2) multi-sequence MRI integration and self-supervised pretraining to improve MRI feature quality, and (3) evaluation on external cohorts to confirm generalisability.

\end{document}